\pdfoutput=1
\documentclass[11pt]{article}
\usepackage[table,xcdraw]{xcolor}
\usepackage{algorithm}
\usepackage{algpseudocode}
\usepackage{amsmath}
\usepackage[preprint]{acl}
\usepackage{amsmath}
\usepackage{pifont}

\usepackage{times}
\usepackage{latexsym}
\usepackage{multirow}
\usepackage{graphicx}
\usepackage[T1]{fontenc}
\usepackage[utf8]{inputenc}

\usepackage{microtype}

\usepackage{inconsolata}

\usepackage{graphicx}

\title{ ChartLens: Fine-grained Visual Attribution in Charts}

\author{Manan Suri $^ {\includegraphics[width=0.35cm]{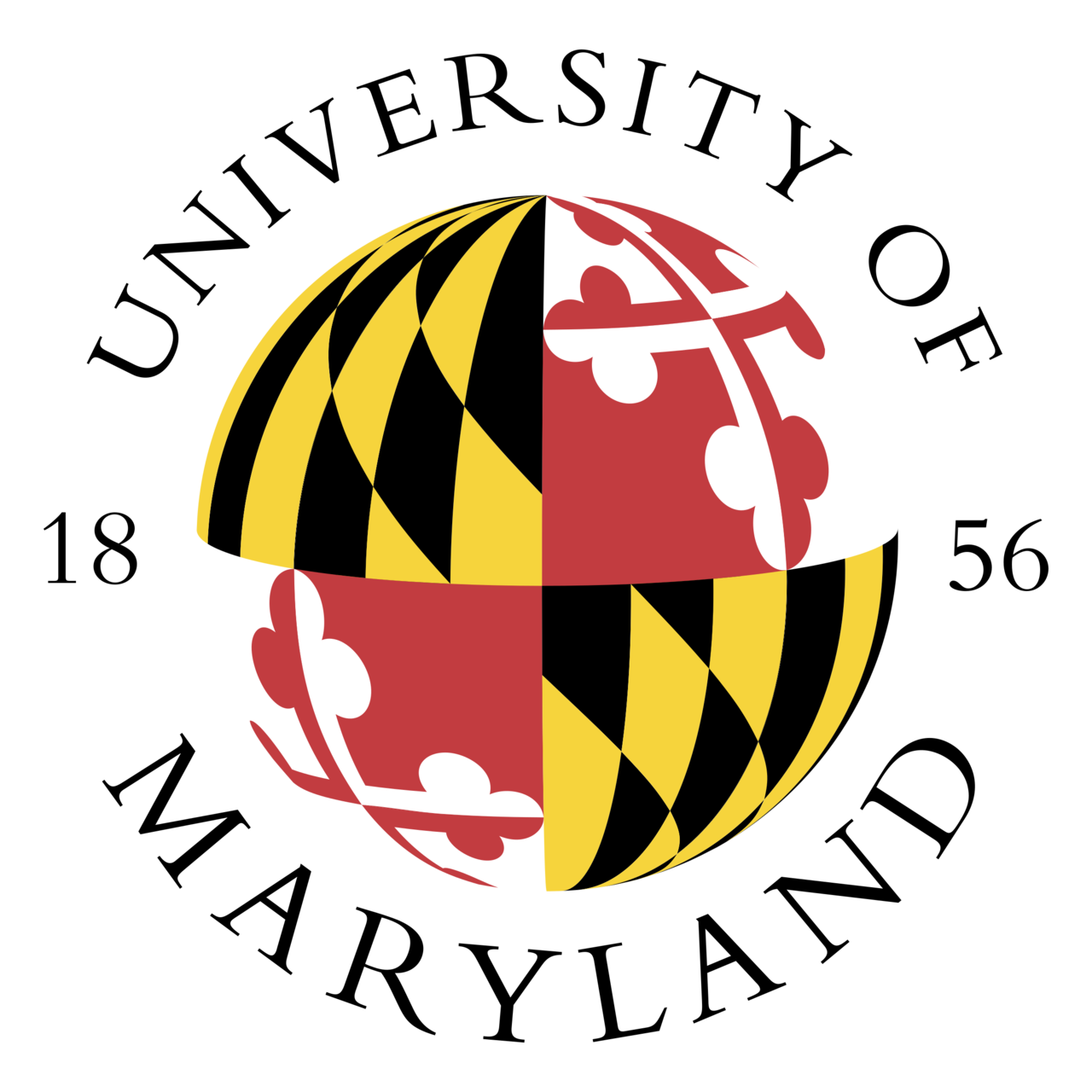}}$, \text{Puneet Mathur} $^ {{\includegraphics[width=0.2cm]{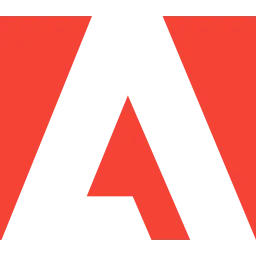}}}$ \thanks{Primary Research Mentor},  \textbf{\text{Nedim Lipka}} $^ {{\includegraphics[width=0.2cm]{figures/722666.png}} }$, \\ \textbf{\text{Franck Dernoncourt}}$ ^ {\includegraphics[width=0.2cm]{figures/722666.png}} $, \hspace{1.5pt} \textbf{\text{Ryan A. Rossi}} $^ {{\includegraphics[width=0.2cm]{figures/722666.png}} }$,\textbf{ \text{Dinesh Manocha}} $^ {\includegraphics[width=0.35cm]{figures/university-of-maryland-logo-1.png}}$\\ $^{\includegraphics[width=0.35cm]{figures/university-of-maryland-logo-1.png}}$ $\text{University of Maryland, College Park}$ \hspace{10pt} $^{\includegraphics[width=0.2cm]{figures/722666.png}} \text{Adobe Research} $\hspace{10pt}   \\ \texttt{manans@umd.edu}, \texttt{puneetm@adobe.com}  }

\begin{document}
\maketitle
\begin{abstract}
The growing capabilities of multimodal large language models (MLLMs) have advanced tasks like chart understanding. However, these models often suffer from hallucinations, where generated text sequences conflict with the provided visual data. To address this, we introduce Post-Hoc Visual Attribution for Charts, which identifies fine-grained chart elements that validate a given chart-associated response. We propose ChartLens, a novel chart attribution algorithm that uses segmentation-based techniques to identify chart objects and employs set-of-marks prompting with MLLMs for fine-grained visual attribution. Additionally, we present ChartVA-Eval, a benchmark with synthetic and real-world charts from diverse domains like finance, policy, and economics, featuring fine-grained attribution annotations. Our evaluations show that ChartLens improves fine-grained attributions by 26-66\%.\footnote{\href{https://github.com/MananSuri27/ChartLens}{Code and data}}

\end{abstract}

\section{Introduction}
Rapid advancements in large language models (LLMs) have revolutionized various natural language processing tasks, including understanding, generation, and reasoning \cite{huang2022towards, yang2024harnessing}. Building on this foundation, multimodal large language models (MLLMs), have extended these capabilities to encompass multimodal tasks like image captioning and visual question answering. However, a critical challenge faced by these models is the prevalence of hallucinations—instances where the model generates content that appears plausible but is factually incorrect or inconsistent \cite{hallu1,hallu2,hallu3}. In MLLMs, this issue is particularly pronounced as cross-modal inconsistencies can emerge, where generated text fails to align with provided visual inputs \cite{huang2024visual, guan2024hallusionbench}. To address this, attribution has emerged as a promising strategy for text-based systems, allowing models to reference external sources, thereby enhancing factual reliability. In the context of LLMs, attribution refers to the ability of a model to provide verifiable evidence, such as references or citations, that supports its generated outputs, thereby enhancing factual reliability and trustworthiness \cite{li2023survey}.  Techniques such as direct generated attribution \cite{peskoff2023credible, sun2022recitation}, post-retrieval answering \cite{ye2023effective, li2023llatrieval}, and post-hoc attribution \cite{huo2023retrieving, chen2023complex} aim to mitigate hallucination by enabling users to trace responses back to their origins. For resolving visual hallucinations specifically, post-generation validation approaches like \cite{zhou2023analyzing,lee2023volcano,yin2023woodpecker} have proven effective in aligning textual outputs with visual evidence, ensuring cross-modal consistency and improving the overall trustworthiness of MLLMs.

\begin{figure}
    \centering
    \includegraphics[width=\linewidth]{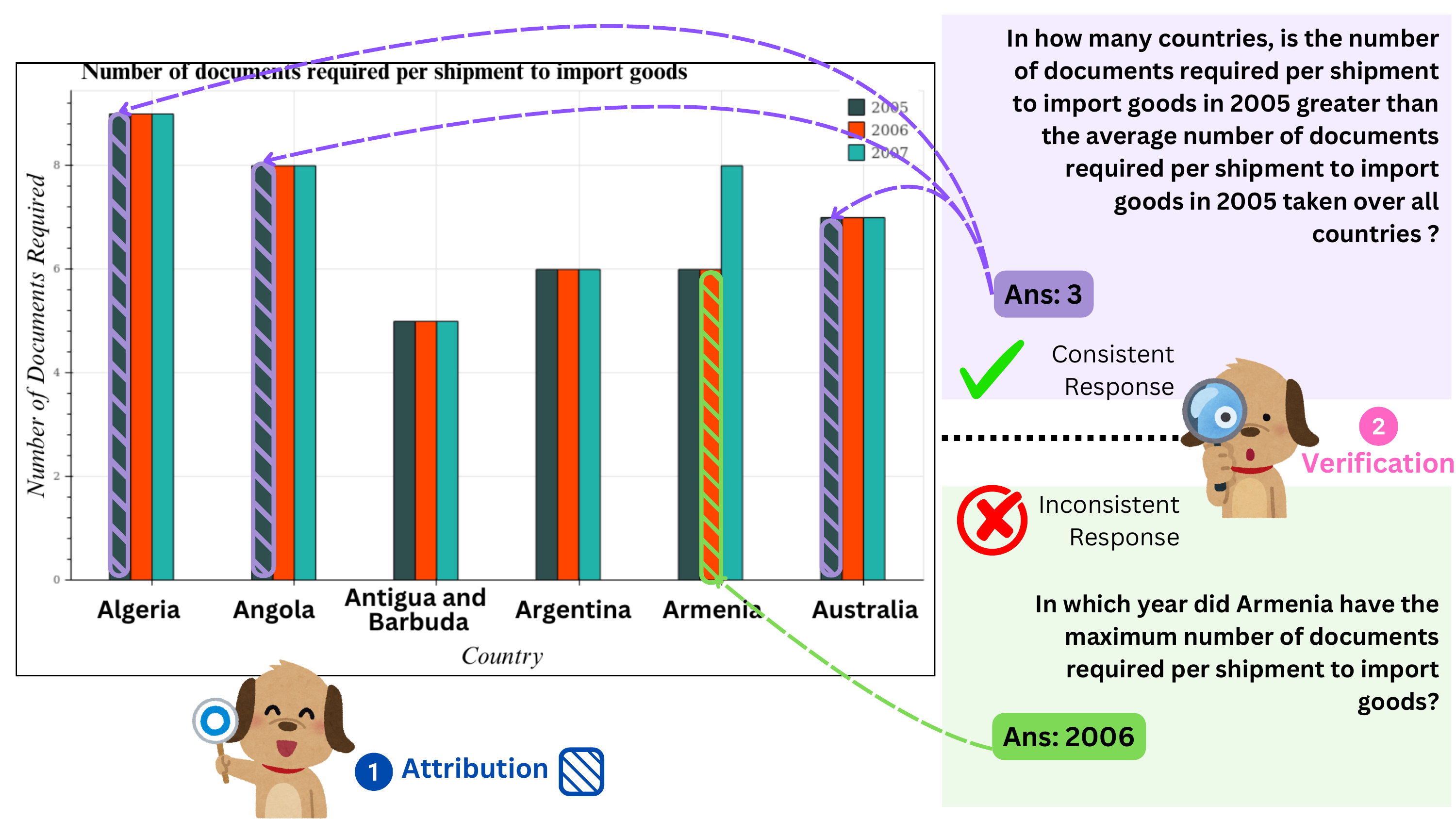}
    \caption{We introduce the task of visual attribution for charts (\textcolor[HTML]{004aad}{\ding{202}}), which grounds textual responses to specific regions in the chart image. This promotes reliable understanding by enabling users to verify claims (\textcolor[HTML]{ff66c4}{\ding{203}}), thus detect potentially hallucinated responses and identifying chart-response misalignments.}
    \label{fig:verification}
\end{figure}

Charts, which are graphical representations of data, play a pivotal role in communicating complex insights across diverse domains, from business analytics to scientific research \cite{embarak2018importance}. As LLMs and multimodal large language models (MLLMs) increasingly handle chart-related tasks—such as chart question answering \cite{kafle2018dvqa,masry2022chartqa,kahou2017figureqa,methani2020plotqa}, captioning \cite{kantharaj2022chart,tang2023vistext,hsu2021scicap}, and chart-to-table \cite{liu2022deplot} conversion—the need for robust validation mechanisms becomes paramount. Unlike textual information, charts encapsulate measurable and exact quantities and often represent intricate relationships like trends, proportions, and comparisons \cite{healy2024data}. Accurately interpreting these relationships requires more than superficial analysis; it demands understanding context-dependent factors like chart type, data encoding, and the layout of visual components such as axes, legends, colors, and shapes. The attribution challenge is further compounded by the need to disentangle overlapping visual elements, resolve ambiguities in labeling, and consistently map visual evidence to textual answers. 

Accurate attribution in chart-related tasks is crucial for ensuring that multimodal large language models (MLLMs) generate reliable and trustworthy outputs \cite{bai2024hallucination}. Charts often convey critical information involving exact quantities, trends, and comparisons. When an MLLM's response to a chart-related request cannot be clearly linked to specific visual elements, it becomes difficult to assess whether the answer is grounded in the chart's data or influenced by hallucinated patterns. This lack of transparency can lead to incorrect conclusions, undermining the reliability of automated systems in critical areas such as financial analysis, policy-making, and scientific research, where accurate data interpretation is essential for decision-making. Reliable attribution helps mitigate these risks by making the model's process verifiable, meaning that the model's response can be traced back to identifiable visual elements in the chart. As demonstrated in Fig~\ref{fig:verification}, this allows for confirmation that the generated response is directly supported by the chart's data, reducing the potential for hallucinated or incorrect interpretations. 

\noindent\textbf{Main Results:} We introduce the task of Post-Hoc Fine-grained Attribution for Charts. This task identifies the specific chart elements (like bars, points, or sectors) that directly support the model's answer to a given question and enable the grounding of model responses to visual elements. We focus specifically on post-hoc attribution since it provides a flexible plug-and-play mechanism agnostic of the actual multimodal chart system used underneath and decouples attribution from response generation for traceability.  

We introduce ChartVA-Eval, a new benchmark designed to advance the evaluation of chart visual attribution systems. ChartVA-Eval comprises real-world chart data sourced from financial documents and policy datasets, such as SEC Filings, the World Bank Open Data, Open Government Data, and the Global Terrorism Database. The benchmark features diverse chart styles and includes retrieval, reasoning, and computation-based questions, all paired with fine-grained visual attribution annotations. We also propose ChartLens, a chart attribution methods that leverages set-of-marks prompting with multimodal LLMs to produce reliable attributions. ChartLens demonstrates significant improvements, achieving 26-66\% higher accuracy compared to competitive baselines, underscoring its effectiveness in identifying the precise chart elements that support model-generated answers.

\textbf{Main Contributions:}

\begin{itemize}
    \item We propose the task of Post-Hoc Fine-grained Visual Attribution for Charts, which focuses on determining the specific chart elements that support a given chart-associated textual response, improving transparency and mitigating hallucination in MLLMs.
    \item We present ChartVA-Eval, a comprehensive benchmark of 1200+ samples, containing real-world chart data from diverse sources. The benchmark features diverse chart styles and fine-grained attribution annotations to facilitate rigorous evaluation.
    \item We introduce ChartLens, a novel chart attribution algorithm based on set-of-marks prompting with multimodal LLMs. Our method achieves 26-66\% improvements over existing baselines, establishing a new state-of-the-art method for chart attribution tasks.
\end{itemize}

\section{Related Work}
\subsection{Response Attribution in LLMs}
Generative LLMs now lead performance in various tasks, but their tendency to produce hallucinations remains a significant challenge \cite{hallu1}. To mitigate these issues, researchers have explored training LLMs to provide citations alongside their answers \cite{gao2023enabling, menick2022teaching, nakano2021webgpt}. Other methods augment LLMs with external tools such as retrievers \cite{ye2023effective, li2023llatrieval}, and search engines \cite{nakano2021webgpt}. 

Three primary attribution strategies have emerged. Direct model-driven attribution allows the model to generate answers and attributions simultaneously, though this often leads to inaccuracies in both the answers and the cited sources \cite{peskoff2023credible, sun2022recitation}. Post-retrieval answering involves explicitly retrieving information first and then answering based on the retrieved data\cite{ye2023effective, li2023llatrieval}. However, retrieval does not always equate to accurate attribution, as conflicts between the model's internal knowledge and the retrieved information can arise \cite{huo2023retrieving, chen2023complex}. In post-generation attribution, the model generates an answer first, and then a search is conducted to find supporting references, modifying the answer if necessary \cite{li2023survey}. 

Additionally, recent research has focused on generating more structured attributions for data from different modalities. For example, MATSA \cite{mathur2024matsa} introduces the Fine-grained Structured Table Attribution (FAST-Tab) task, where a multi-agent LLM system provides row- and column-level attributions to visually support claims derived from tables.

\subsection{Visual Chart Understanding}

Automated chart understanding has seen significant advancements through classification-based and generation-based methods. Early classification models like IMG+QUES \cite{kafle2018dvqa} and Relation Networks \cite{santoro2017simple} faced out-of-vocabulary (OOV) challenges, which were mitigated by dynamic encoding techniques such as SANDY \cite{kafle2018dvqa}  and PReFIL \cite{kafle2020answering} that incorporated OCR sub-networks . Pre-trained models like STL-CQA \cite{singh2020stl} and VisionTaPas \cite{masry2022chartqa} further improved performance by leveraging transformer-based architectures .

Generation-based approaches dominate tasks like chart captioning and chart-to-table conversion. Models such as Donut~\cite{kim2022ocr} and Pix2Struct~\cite{lee2023pix2struct} introduced end-to-end OCR-free architectures, while UniChart \cite{masry2023unichart} and MatCha \cite{liu2022matcha} incorporated chart-specific pre-training objectives.

The emergence of Multimodal Large Language Models (MLLMs) like ChartLlama \cite{han2023chartllama} and ChartAssistant \cite{meng2024chartassisstant} has enabled strong zero-shot performance through instruction-tuning on chart-specific datasets. Additionally, tool-augmented methods such as DePlot \cite{liu2022deplot} and StructChart \cite{xia2023structchart} aid LLMs by converting charts to structured data tables. Despite these advances, challenges remain in handling domain-specific chart diversity and developing robust evaluation metrics \cite{wang2024charxiv}.

\section{Post-Hoc Visual Attribution in Charts} 

\textbf{Problem Statement.}
Given a dataset $ \mathcal{D} $ consisting of a set of charts $ \mathcal{C} $, each chart is an image $ c \in \mathcal{C}, c = \mathcal{I}^{w\times h\times3} $, and is associated associated with a set of responses $ \mathcal{R}_c $. Each response is denoted by $v, v \in \mathcal{R}_c$.

The objective is to determine the set of visual regions within the chart $ c $ that provide evidence for the response $ v $. Specifically, for a given chart $ c $ and response $ v $, the goal is to produce an attribution set $ \mathcal{A}_{c,v} $, where:  

\begin{equation}
\mathcal{A}_{c,v} = \{ a_1, a_2, \ldots, a_n \}
\end{equation}  

and each $ a_i $ represents a distinct region corresponding to an element in the chart $ c $ (e.g., bars, lines, points, segments) that supports the response $ v $.  

The expectation is that $ \mathcal{A}_{c,v} $ satisfies the following criteria:  

\begin{enumerate}
    \item \textbf{Relevance}: Each region $ a_i $ must be directly relevant to the response $ v $.  
    \item \textbf{Completeness}: The set $ \mathcal{A}_{c,v} $ should comprehensively cover all the visual evidence needed to justify $ v $.  
    \item \textbf{Precision}: The regions $ a_i $ should be specific and exclude irrelevant parts of the chart.  
\end{enumerate}  

The task can be summarized as finding a mapping function:  

\begin{equation}
f: (c, v) \mapsto \mathcal{A}_{c,v}
\end{equation}  

where $ f $ identifies the precise visual elements in $ c $ that substantiate the response $ v $.

\section{ChartVA-Eval}
In this section, we introduce ChartVA-Eval, a benchmark designed to evaluate visual attribution in charts. Each data point in ChartVA-Eval consists of a chart image $c \in \mathcal{C}$, represented as $c = \mathcal{I}^{w \times h \times 3}$, a textual response $v \in \mathcal{R}c$, and a ground truth attribution set $\mathcal{A}{c,v}^{gt}$, which represents the ground truth attributions. Additionally, the set of all regions in the chart corresponding to different elements is denoted as $\mathcal{A}_c$, representing all potential regions within the chart.

\begin{table}[]
\resizebox{\columnwidth}{!}{%
\begin{tabular}{cccc}
\hline
\textbf{Dataset}                       & \textbf{ChartVA- AITQA} & \textbf{ChartVA- PlotQA} & \textbf{ChartVA-ChartQA} \\ \hline
\multicolumn{1}{c|}{\# of Queries}     & 301                     & 595                      & 348                      \\
\multicolumn{1}{c|}{\# of Charts}      & 301                     & 581                      & 266                      \\
\multicolumn{1}{c|}{\# of Bar Charts}  & 203                     & 396                      & 121                      \\
\multicolumn{1}{c|}{\# of Pie Charts}  & 0                      & 0                        & 109                      \\
\multicolumn{1}{c|}{\# of Line Charts} & 98                     & 199                      & 118                      \\
% \multicolumn{1}{c|}{Domain}  &  Finance, Corporate        &    Economics, Governance                 &         Demographics, Global Development                                         \\
\multicolumn{1}{c|}{Chart Source}      & Synthetic               & Synthetic                & Real World                    \\
\multicolumn{1}{c|}{Multiple Attributions} & No                  & Yes                      & Yes                      \\
\multicolumn{1}{c|}{Avg \# of Attributions} & 1                  & 2.4                      & 1.43                     \\
\multicolumn{1}{c|}{Max \# of Attributions} & 1                  & 12                       & 8                        \\
\multicolumn{1}{c|}{Avg \# of Data Series}  & 1.23               & 2.52                     & 2.45                     \\
\multicolumn{1}{c|}{Max \# of Data Series}  & 8                  & 4                        & 14                       \\ \hline
\end{tabular}%
}
\caption{Statistics for the ChartVA-Eval benchmark, reported for constituent datasets.}
\label{tab:dataset}
\end{table}

\subsection{Data Sources}
The ChartVA-Eval Benchmark is constructed from a diverse set of data sources to ensure comprehensive evaluation of post hoc attribution in chart-based visual question answering (VQA). By incorporating both synthetic and real-world charts, it captures a wide range of design styles, chart types, and question-answer (QA) contexts. This diversity enables rigorous assessment of model performance across different domains and visual complexities. The benchmark draws from three key datasets: MATSA-AITQA \cite{mathur2024matsa}, PlotQA \cite{methani2020plotqa}, and ChartQA \cite{masry2022chartqa}, each offering unique characteristics and challenges.

MATSA-AITQA \cite{mathur2024matsa} provides chart data derived from tabular QA over public U.S. SEC filings of major airline companies, covering the fiscal years 2017 to 2019 \cite{katsis2021aitqaquestionansweringdataset}. The tables are paired with QA pairs and annotated cells corresponding to the data points supporting the answers. From these tables, synthetic charts are generated by applying variations in themes, color palettes, fonts, and design elements like grid lines and tick styles, resulting in over $O(10^4)$ possible style combinations. Each QA pair is associated with a single visual attribution. The dataset includes chart types such as grouped bar charts, stacked bar charts, simple bar charts (both horizontal and vertical), line charts. Further details for synthetic chart generation are present in the appendix. PlotQA \cite{methani2020plotqa} focuses on synthetic scientific charts paired with bounding box annotations and diverse reasoning-based questions. The dataset includes line charts and bar charts (both vertical and horizontal), with one or more visual elements annotated to support each answer. The data for these charts is sourced from publicly available repositories, including the World Bank Open Data, Open Government Data, and the Global Terrorism Database. This controlled synthetic environment allows for evaluating fine-grained attribution tasks that require careful interpretation and logical reasoning. ChartQA \cite{masry2022chartqa} offers real-world charts accompanied by human-authored QA annotations. The charts are sourced from platforms such as Statista, Pew Research Center, Our World in Data (OWID), and the Organisation for Economic Co-operation and Development (OECD). The dataset includes a variety of chart types, particularly pie charts, line charts, and bar charts. Given the scarcity of pie charts in other datasets, they are oversampled to ensure balanced representation. ChartQA captures the complexities and variability found in real-world data visualizations, providing a realistic benchmark for evaluating attribution models.

\begin{figure*}
    \centering
    \includegraphics[width=\linewidth]{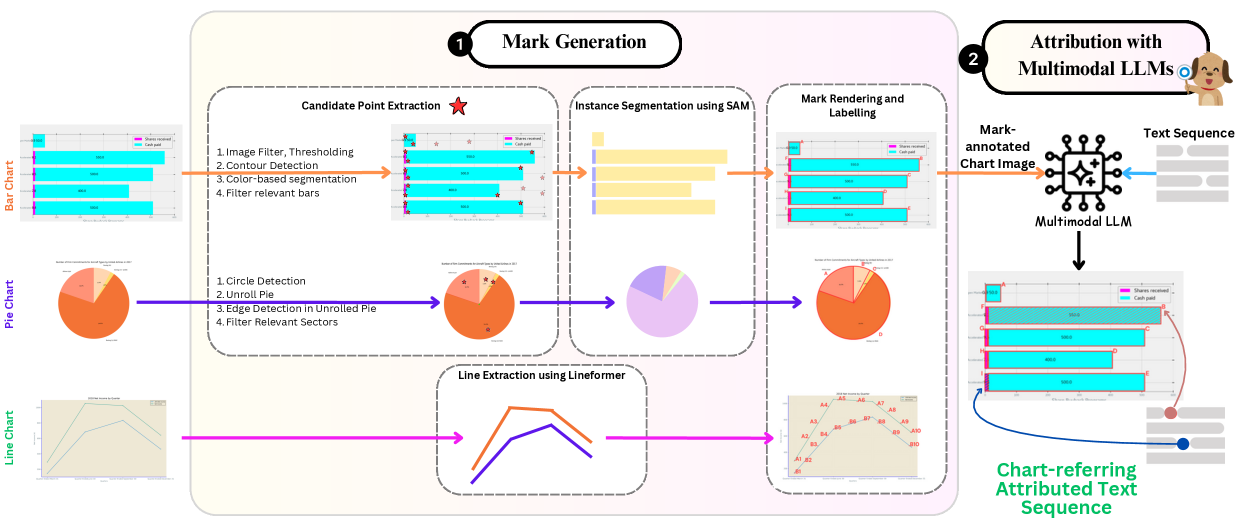}
    \caption{\textbf{ChartLens:} \ding{202} Chart elements, such as bars and pie sectors, are extracted through heuristic-guided methods and refined using SAM, while lines are segmented using Lineformer. \ding{203} The segmented elements are then marked, labeled, and used to prompt multimodal LLMs, enabling fine-grained attribution by grounding textual responses to visual regions.}
    \label{fig:chartlens}
\end{figure*}

\subsection{Attribution Annotation}
For the ChartQA and PlotQA datasets, we employed a hybrid approach to generate attribution annotations, combining large-scale automated annotation with human verification. We utilized GPT-4o to generate initial annotations by leveraging the underlying data tables, questions, and answers. Specifically, we identified frequent question templates and designed tailored prompts for each template. For example, for cardinality related QA pairs, the model was instructed to select all data points counted in the cardinality. These automated annotations were subsequently refined through human validation. In an interactive setting, annotators reviewed the rendered bounding boxes on the charts and assessed the annotations based on two criteria: (1) Relevance — ensuring the annotated elements directly support the answer, and (2) Completeness — verifying that all necessary chart elements were included. This process ensured high-quality and precise attribution annotations for both datasets. Further details on attribution annotation are provided in the appendix.

\section{ChartLens}
ChartLens (Fig~\ref{fig:chartlens}) facilitates fine-grained visual attribution for charts by first detecting and labeling chart elements with distinct marks. These marks, which serve as referable visual anchors, are then used to prompt MLLMs to attribute responses to specific visual elements within the chart.

\subsection{Mark Generation}

The goal of mark generation is to identify and tag fine-grained visual features within chart images to form a set of candidates for attribution. These marks serve as visual anchors to prompt multimodal LLMs by providing locality-based grounding. Effective mark generation requires the ability to isolate individual chart components, while ensuring robustness across various chart types and visual styles.

\noindent\textbf{Heuristic-guided Instance Segmentation} 
For segmenting bar charts, the input image is first binarized using Otsu thresholding applied to both the RGB and HSV representations. If the chart has a dark background, the binarized image is inverted to ensure foreground features, such as bars, are correctly highlighted. From the binarized image, an initial set of contours is generated. These contours are further refined by breaking them down using unique pixel values, isolating individual bars. To eliminate irrelevant or spurious contours, a filtering step based on solidity and area thresholds is applied, ensuring that only well-defined bars are retained.

For pie charts, segmentation begins by identifying the largest contour in the binarized image, which typically corresponds to the pie chart itself. We compute the minimum enclosing circle for this contour to approximate the chart's boundary. Following the approach in \cite{savva2011revision}, the pie chart is unrolled along the radial axis to create a linear representation. In this unrolled form, complete edges are detected to identify sector boundaries, which are then mapped back to the original circular region. This process yields segments corresponding to individual slices of the pie chart.

While these heuristic methods leverage the structural and geometric properties of charts effectively, they suffer from several limitations. They are sensitive to noise and perform poorly on low-contrast images, often misidentifying irrelevant components such as grid lines or labels as chart elements. To address these issues, we employ the Segment Anything Model (SAM) \cite{kirillov2023segment} for instance segmentation. Specifically, $n$ points are sampled from each detected element and used as prompts for SAM. The model generates masks that accurately enclose the objects associated with the sampled points, overcoming the shortcomings of classical methods.

SAM's architecture allows it to handle noisy and low-quality images more robustly. It produces precise masks that closely align with the boundaries of chart elements, even in complex cases. Additionally, SAM naturally suppresses background features like grid lines by generating weaker masks (low IoU) for these elements, as they lack the spatial coherence of primary chart components. Unlike heuristic approaches, SAM generalizes well across diverse chart types and layouts without requiring extensive parameter tuning. By combining heuristic-guided preprocessing with SAM-based instance segmentation, we achieve a more flexible and accurate segmentation process that leverages the strengths of both classical computer vision and modern deep learning techniques.

\noindent\textbf{Transformer-based Line Segmentation} 
We use LineFormer \cite{lal2023lineformer} to extract lines from line charts. Lines present unique challenges for segmentation due to their fine structural features, such as narrow width, overlapping trajectories, and the presence of intersecting lines. These characteristics make it difficult for classical computer vision methods or point-based prompting approaches to accurately identify and segment lines, especially in dense or complex charts.

LineFormer effectively addresses these challenges. It leverages the global context provided by the transformer architecture to distinguish lines even when they are closely spaced or intersecting. After detecting candidate lines with LineFormer, we divide each line into equally spaced segments along its domain extent (horizontal range). These smaller segments serve as fine-grained marks for our attribution algorithm.

\subsection{Attribution with MLLMs}

To facilitate accurate attribution in chart-based tasks, we employ \textit{Set-of-Marks (SoM) prompting}, a visual prompting technique designed to leverage the visual grounding capabilities of multimodal LLms. Inspired by \cite{yang2023set}, SoM prompting partitions an image into regions of varying granularity using interactive segmentation models like SEEM or SAM. These segmented regions are then overlaid with visual marks, such as alphanumeric labels, masks, or bounding boxes. This marked image is presented as input to the multimodal LLM. SoM prompting is effective because it enables explicit localization within the image, helping the model isolate distinct regions and understand their spatial relationships. Additionally, by labeling these elements, the technique simplifies reasoning for the model, making it easier to reference specific components during visual grounding tasks. The combination of these factors enhances the model’s ability to interpret and connect visual information with textual queries.

In our approach, we prompt multimodal LLMs with chart images overlaid with marks. The prompts are structured to achieve two primary goals: validation and attribution. The prompt first explains the concept of chart attribution, providing a few-shot set of textual examples of question-answer (QA) pairs along with their corresponding attribution. Next, the model is instructed to follow a chain-of-thought (CoT) reasoning process to perform step-wise validation and attribution.

\noindent\textbf{Validation} involves verifying whether the QA pair is consistent with the information in the chart image. The model evaluates if the answer aligns with the visual elements and data presented in the chart.

\noindent\textbf{Attribution} requires the model to identify and mention the specific labeled elements within the chart that support the given answer. By explicitly referencing these elements, the model's response becomes more transparent and easier to verify.

% This combination of SoM prompting, few-shot examples, and CoT-based reasoning enhances the model's ability to ground its answers in the chart data, thereby improving the accuracy and reliability of chart attribution tasks.

\begin{table*}[h!]
\centering
\resizebox{1.4\columnwidth}{!}{%
\begin{tabular}{c|ccc|ccc|ccc}
\hline
\multirow{2}{*}{\textbf{Baseline}} & \multicolumn{3}{c|}{\textbf{ChartVA - AITQA}} & \multicolumn{3}{c|}{\textbf{ChartVA - PlotQA}} & \multicolumn{3}{c}{\textbf{ChartVA - ChartQA}} \\ \cline{2-10} 
 & \textbf{P} & \textbf{R} & \textbf{F1} & \textbf{P} & \textbf{R} & \textbf{F1} & \textbf{P} & \textbf{R} & \textbf{F1} \\ \hline
\textbf{Zero-shot ChatGPT4o} & 22.33 & 23.23 & 22.77 & 3.40 & 3.21 & 3.30 & 8.13 & 7.41 & 7.75 \\
\textbf{KOSMOS2} & 0.51 & 0.51 & 0.51 & 1.60 & 0.74 & 1.01 & 3.31 & 2.96 & 3.13 \\
\textbf{LISA} & 0.83 & 29.29 & 1.62 & 0.18 & 6.39 & 0.34 & 0.52 & 30.37 & 1.01 \\ \hline
\textbf{ChartLens} & \textbf{79.86} &\textbf{ 61.17} &\textbf{ 69.28} & \textbf{35.38} & \textbf{33.94} & \textbf{34.65} & \textbf{74.51}	& \textbf{56.30}	& \textbf{64.14} \\ \hline
\end{tabular}%
}
\caption{Comparison of ChartLens with baselines on the ChartVA-Eval benchmark for bar charts.}
\label{tab:bar}
\end{table*}

\begin{table*}[h!]
\centering
\resizebox{1.9\columnwidth}{!}{%
\begin{tabular}{c|cc|cc|cc}
\hline
\multirow{2}{*}{\textbf{Baseline}} & \multicolumn{2}{c|}{\textbf{ChartVA - AITQA}} & \multicolumn{2}{c|}{\textbf{ChartVA - PlotQA}} & \multicolumn{2}{c}{\textbf{ChartVA - ChartQA}} \\ \cline{2-7} 
 & \textbf{Detection \% ($\uparrow$)} & \textbf{Chart Ar \% ($\downarrow$)} & \textbf{Detection \% ($\uparrow$)} & \textbf{Chart Ar \% ($\downarrow$)} & \textbf{Detection \% ($\uparrow$)} & \textbf{Chart Ar \% ($\downarrow$)} \\ \hline
\textbf{Zero-shot ChatGPT4o} & 18.28 & 1.94 & 6.79 & \textbf{8.63} & 3.39 & \textbf{1.15} \\
\textbf{KOSMOS2} & 74.19 & 46.03 & 38.83 & 27.06 & 87.29 & 41.49 \\
\textbf{LISA} & \textbf{94.62} & 63.18 & 50.21 & 40.92 & 50.21 & 40.92 \\ \hline
\textbf{ChartLens} & 59.14 & 	\textbf{1.25} & 	\textbf{51.84} & 	9.98 & 	\textbf{77.8} & 	5.34  \\ \hline
\end{tabular}%
}
\caption{Comparison of ChartLens with baselines on the ChartVA-Eval benchmark for line charts.}
\label{tab:line}
\end{table*}

\begin{table}[h!]
\centering
\resizebox{0.7\columnwidth}{!}{%
\begin{tabular}{c|ccc}
\hline
\multirow{2}{*}{\textbf{Baseline}} & \multicolumn{3}{c}{\textbf{ChartVA - ChartQA}} \\ \cline{2-4} 
 & \textbf{P} & \textbf{R} & \textbf{F1} \\ \hline
\textbf{Zero-shot ChatGPT4o} & 8.94 & 5.99 & 7.17 \\
\textbf{KOSMOS2} & 20.18 & 8.24 & 11.70 \\
\textbf{LISA} & 1.32 & 13.86 & 2.41 \\ \hline
\textbf{ChartLens} &\textbf{53.33}	&\textbf{44.57}	&\textbf{48.56} \\ \hline
\end{tabular}%
}
\caption{Comparison of ChartLens with baselines on the ChartVA-Eval benchmark for pie charts.}
\label{tab:pie}
\end{table}

\section{Experiments}

\subsection{Baselines}

\textbf{Zero-shot GPT-4o Bounding Box Prompting:} As a baseline, we prompt GPT-4o \cite{openai_gpt4o_2024} to predict normalized bounding box coordinates for chart components (e.g., lines, bars, pie sectors) based on input text and the visual chart. This approach aligns with prior work for zero-shot localization tasks.

\noindent\textbf{Kosmos-2:} Kosmos-2 \cite{peng2023kosmos} is a multimodal large language model (MLLM) trained on grounded image-text data (GrIT) that integrates text-to-visual grounding capabilities. By representing object locations as Markdown links, it enables tasks such as referring expression comprehension, phrase grounding, and multimodal reasoning, and generates bounding boxes for visual grounding tasks.

\noindent\textbf{LISA:} LISA (Large Language Instructed Segmentation Assistant) \cite{li2023llatrieval} is a reasoning-based segmentation model that generates masks from implicit and complex textual queries. By introducing a <SEG> token and leveraging the embedding-as-mask paradigm, LISA extends MLLM capabilities to reasoning segmentation with robust zero-shot performance and further improves with minimal task-specific fine-tuning.

\subsection{Evaluation}  

\textbf{Bar Charts and Pie Charts:} Detected regions are first matched to ground truth regions (e.g., bars or sectors in the chart) based on a threshold Intersection over Union (IoU) value of $ \text{IoU} \geq 0.9 $. The matched regions are treated as discrete items, where each detected region is assigned a unique label corresponding to its ground truth region. The performance is evaluated using Precision, Recall, and F1-score, computed over the set of filtered detected regions and ground truth regions. Let $ D $ denote the set of detected regions after filtering, and $ G $ denote the set of ground truth regions. Precision ($ P $), Recall ($ R $), and F1-score ($ F1 $) are defined as $ 
P = \frac{|D \cap G|}{|D|}, \quad R = \frac{|D \cap G|}{|G|}, \quad F1 = \frac{2 \cdot P \cdot R}{P + R}$.

\noindent\textbf{Line Charts:} Unlike bar charts and pie charts, where detected regions can be matched to discrete ground truth regions, the task for line charts involves referring to singular points. Since grounding models generate bounding boxes or regions, it is challenging to precisely match these regions to individual ground truth points without ambiguity. To address this, two metrics are defined for evaluation:
\begin{enumerate}  
        \item \textbf{Detection Rate:} Measures the proportion of ground truth points covered within the detected region(s), analogous to recall.  
        \item \textbf{Average Area Detected:} Large detected areas indicate low precision, even if the recall is high. This is quantified as the average percentage of the input chart image covered. Larger detected areas result in higher recall but lower precision, making this metric critical for evaluating the trade-off between precision and recall in line chart attribution tasks.  
\end{enumerate}

For ChartLens, Large Multimodal Models (LMMs) are prompted to detect pairs of marked points on the line between which the attribution lies. These point pairs are considered corners of a bounding box, and the same metrics are used.

\subsection{Implementation Details}
\begin{figure*}[h!]
\centering
    \includegraphics[width=\linewidth]{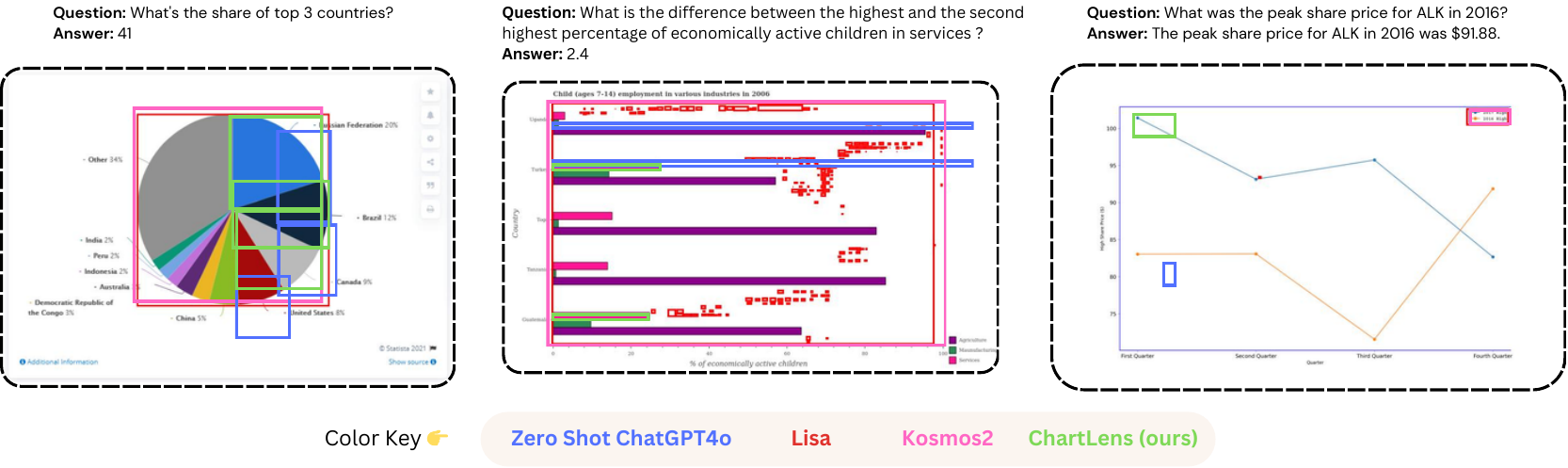}
    \caption{Qualitative comparison of our ChartLens with the baselines. ChartLens is able to effectively localize relevant, complete and precise attributions in the chart images.}
    \label{fig:qual}
\end{figure*}
The base multimodal language model (MLLM) for ChartLens is \texttt{ChatGPT-4o}, which is used for zero-shot bounding box detection and attribution tasks. For LISA, we use the \texttt{xinlai/LISA-13B-llama2-v1} checkpoint with 8-bit quantization and mixed-precision (\texttt{fp16}) to optimize memory efficiency during inference. Kosmos-2 is implemented using the \texttt{microsoft/kosmos-2-patch14-22} checkpoint. We use the \texttt{facebook/sam-vit-large} checkpoint for SAM. All experiments are conducted using a single NVIDIA A5000, over 6 hours. Default hyperparameters from each model's implementation are used unless stated otherwise.

\section{Results}
In this section, we present a detailed comparison of the proposed ChartLens model against several baselines, including Zero-shot ChatGPT4o, KOSMOS2, and LISA, across three different chart types: bar charts, line charts, and pie charts. The results demonstrate that ChartLens consistently outperforms the baselines across all chart types, highlighting its robustness and effectiveness in visual chart understanding.

\noindent\textbf{Bar Charts}
ChartLens demonstrates significant performance improvements over all baselines in bar charts, achieving F1 scores of 69.28 on ChartVA-AITQA, 34.65 on ChartVA-PlotQA, and 64.14 on ChartVA-ChartQA (Table~\ref{tab:bar}). In contrast, Zero-shot ChatGPT4o achieves much lower F1 scores of 22.77, 3.30, and 7.75, reflecting its limitations in numerical reasoning and visual attribution. KOSMOS2 and LISA perform poorly, with F1 scores below 5 across benchmarks, highlighting their inability to handle bar charts due to insufficient grounding of visual and numerical reasoning.

\noindent\textbf{Line Charts}
For line charts (Table~\ref{tab:line}), ChartLens achieves strong detection accuracy of 59.14\%, 51.84\%, and 77.8\% on ChartVA-AITQA, PlotQA, and ChartQA, respectively, with low chart area errors of 1.25\%, 9.98\%, and 5.34\%. While LISA and KOSMOS2 achieve high detection rate, this can largely be explained by the high Chart\% area covered by their attributions; covering large areas of the chart makes capturing specific points non-trivial but reduces the specificity of attributions, making them less effective at fine-grained localization. In contrast, ChartLens reduces Chart\% area by $\approx$ 3-50 times.

\noindent\textbf{Pie Charts}
ChartLens outperforms baselines in pie charts, achieving an F1 score of 48.56, significantly higher than Zero-shot ChatGPT4o (7.17), KOSMOS2 (11.70), and LISA (2.41) (Table~\ref{tab:pie}). Its precision (53.33) and recall (44.57) confirm its ability to attribute pie chart segments accurately. In contrast, Zero-shot ChatGPT4o and KOSMOS2 struggle with interpreting proportions, while LISA's extremely low performance highlights its difficulty in handling pie chart geometry and segmentation tasks.

\noindent Figure~\ref{fig:qual} shows a qualitative comparison of ChartLens with the baselines across bar charts, line charts, and pie charts. ChartLens consistently identifies and attributes relevant chart elements more accurately than the baselines, demonstrating a clear understanding of numerical and visual relationships. Zero-shot ChatGPT4o attempts to make fine-grained specific selections, however fails to exhibit robust localization since it expresses attributions using text based coordinates. LISA and KOSMOS2 consistently refer to typical chart components, like the pie as a whole, or the entire area but do not exhibit sensitivity to given queries.  

\section{Conclusion and Future Work}
In this work, we introduced the task of Post-Hoc Fine-grained Visual Attribution for Charts, addressing the challenge of grounding chart-related responses to specific visual elements. To facilitate this, we proposed ChartLens, a novel attribution algorithm leveraging segmentation techniques and set-of-marks prompting with multimodal LLMs. Additionally, we presented ChartVA-Eval, a comprehensive benchmark featuring real-world and synthetic charts across diverse domains, enabling rigorous evaluation of visual attribution methods. Our experiments demonstrated that ChartLens significantly outperforms competitive baselines, achieving improvements of 26-66\%. By enhancing transparency and mitigating hallucinations in MLLMs, our work lays a foundation for reliable chart interpretation in critical applications such as financial analysis, policy-making, and scientific research. Future work will explore extending these methods to other forms of visual data and improving robustness across chart styles and complexities.

\section{Ethics Statement}
This research utilizes publicly available datasets, ensuring compliance with their respective licenses. The identities of human evaluators remain confidential, and no personally identifiable information (PII) is used at any stage of our experiments. Our work is designed specifically for fine-grained visual attribution applications and does not extend to other use cases. We acknowledge the broader challenges associated with large language models (LLMs), including potential risks related to misuse and safety, and encourage readers to consult relevant literature for a detailed exploration of these issues \citep{risks1, risks2, risks3}.

\section{Limitations}
While our work makes significant strides in fine-grained visual attribution for charts, it has certain limitations. 

First, the system relies on segmentation as a core component, and any inaccuracies in the segmentation process may result in imperfect or incomplete attributions. However, as segmentation is modular, it can be improved or replaced with more advanced methods in future iterations. 

Second, our approach primarily focuses on visual chart elements, such as bars, points, or sectors, and does not account for textual components like captions, labels, or titles. Addressing this limitation and integrating text-based reasoning alongside visual attribution remains a promising direction for future research.
\bibliography{custom}

\appendix

\section{Details on Benchmark Construction}
\label{sec:appendix}
\subsection{Datasets}
\subsubsection{MATSA}
The TabCite dataset from MATSA \cite{mathur2024matsa},consists of tables derived from various sources such as Wikipedia pages and SEC filings. The TabCite benchmark is built by reformulating existing datasets like TOTTO \cite{parikh2020totto}, FetaQA \cite{nan2022fetaqa}, and AITQA \cite{katsis2021aitqaquestionansweringdataset} to create QA pairs with human-generated questions, free-form answers, and ground truth row/column attributions. The dataset focuses on fine-grained table structure attribution, particularly identifying rows and columns for accurate reasoning and table-based question answering. MATSA, in comparison to other models, performs well across multiple settings, achieving the best F1 scores for fine-grained attribution, indicating its effectiveness for reasoning over tables with complex structures.

For ChartVA-Eval, we selected the AITQA subdataset due to its simpler table structure, which allowed for easier and more traceable conversion from tables to charts without losing attribution. Additionally, AITQA is the only subdataset that contains numerical values in every cell, making it ideal for generating charts that can be directly derived from the data. The numerical consistency ensures that the table-to-chart conversion maintains the integrity of the data, allowing for accurate visualization and reasoning over the tables.

\subsubsection{PlotQA}
PlotQA \cite{methani2020plotqa} is a dataset designed for the task of reasoning over real-world plots. It includes data sourced from various online platforms like World Bank Open Data and the Global Terrorism Database, covering a wide range of indicator variables such as fertility rates, rainfall, and coal production across different years, countries, and regions. The dataset comprises 841 unique variables and 160 entities, with data spanning from 1960 to 2016. These statistics are represented in three main plot types: bar plots, line plots, and scatter plots. The plots vary in their visual elements, including legend positions, fonts, grid lines, and color schemes, allowing for rich and diverse plot representations. In total, 224,377 unique plots were generated, ensuring a comprehensive coverage of data.

To facilitate the creation of complex reasoning tasks, the PlotQA dataset also features a collection of 7,000 crowd-sourced questions, which were generated by workers on Amazon Mechanical Turk. These questions were categorized into three types: structural understanding, data retrieval, and reasoning. By analyzing the crowd-sourced questions, the authors extracted 74 question templates that were manually paraphrased to ensure natural phrasing. This process aimed to ensure that the dataset more accurately reflects real-world challenges in plot interpretation, providing a rich resource for training and evaluating machine learning models for visual reasoning tasks. The resulting dataset is notable for its realistic question vocabulary, longer questions, and diverse set of answers, making it a significant step forward in the field of plot-based question answering.
\subsubsection{ChartQA}
ChartQA \cite{masry2022chartqa} is a benchmark dataset designed to evaluate question answering (QA) models over chart images. It consists of a diverse collection of charts crawled from four sources: Statista, Pew Research, Our World In Data (OWID), and the OECD. These charts cover various topics such as economics, politics, and global issues, and include bar, line, and pie charts. To enhance the dataset's coverage, two main methods of annotation were employed: human-authored QA pairs collected via Amazon Mechanical Turk (AMT) and machine-generated questions derived from human-written chart summaries. The dataset focuses on two types of questions: compositional (involving logical or mathematical operations) and visual (related to chart attributes like color or height), which are designed to test complex reasoning abilities.

The dataset also employs data augmentation through the fine-tuning of a T5 model on SQuAD to generate diverse, human-like questions from chart summaries. This process helps to introduce linguistic variations and enriches the dataset with syntactic complexity. A significant feature of ChartQA is its coverage of both simple and complex charts, with the latter including multi-column charts like stacked bars and multi-line graphs. With 6,150 unique tokens in questions and 4,319 in answers, ChartQA presents a challenging task for QA models, reflecting real-world scenarios where questions require intricate reasoning over chart data. The dataset's broad topic coverage and diverse question types make it an essential resource for advancing research in visual question answering and complex reasoning over data visualizations.
\subsection{Human Annotation}  
We employed three graduate student annotators, aged 23-26, with prior experience working with charts across various domains. The annotators were fairly compensated at the standard Graduate Assistant hourly rate, following their respective graduate school policies.  

\begin{figure*}
    \centering
    \includegraphics[width=0.8\linewidth]{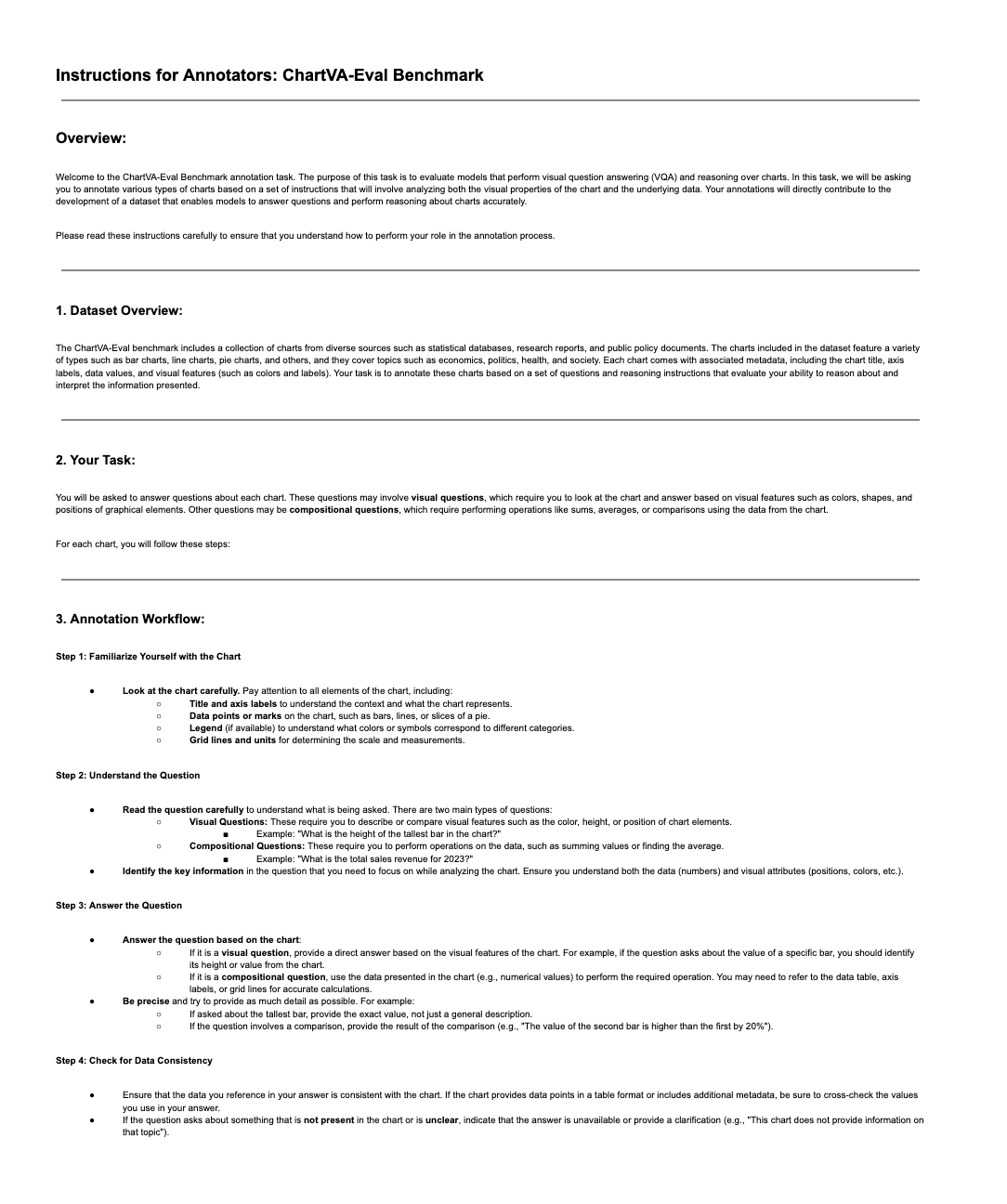}
    \caption{Overview of annotation guidelines provided to annotators for ensuring accurate and consistent visual attributions.}
    \label{fig:annotation}
\end{figure*}

The purpose of the annotation process was to ensure high-quality and precise visual attributions by refining automated annotations and verifying their correctness. Specifically, the annotators were tasked with reviewing bounding box annotations to assess Relevance, ensuring that the annotated chart elements directly supported the provided answers, and Completeness, verifying that all necessary chart elements were included. The annotation process was conducted in an interactive setting where annotators could inspect rendered visualizations and iteratively refine the annotations as required. An overview of the annotation instructions can be seen in Fig \ref{fig:annotation}.  

To evaluate the consistency and reliability of the annotations, we calculated inter-annotator agreement metrics using Cohen's Kappa (\(\kappa\)). For Relevance, the overall agreement was near perfect with a Kappa score of 0.89. Similarly, for Completeness, the agreement remained strong, achieving a Kappa score of 0.84. Additionally, pairwise Kappa scores between the three annotators were computed to further validate consistency: Annotator 1 and Annotator 2 achieved a \(\kappa\) of 0.87, Annotator 1 and Annotator 3 reported a \(\kappa\) of 0.85, while Annotator 2 and Annotator 3 achieved a \(\kappa\) of 0.83.  

\subsection{Synthetic Chart Construction}
\begin{figure*}[h!]
    \centering
    \includegraphics[width=\linewidth]{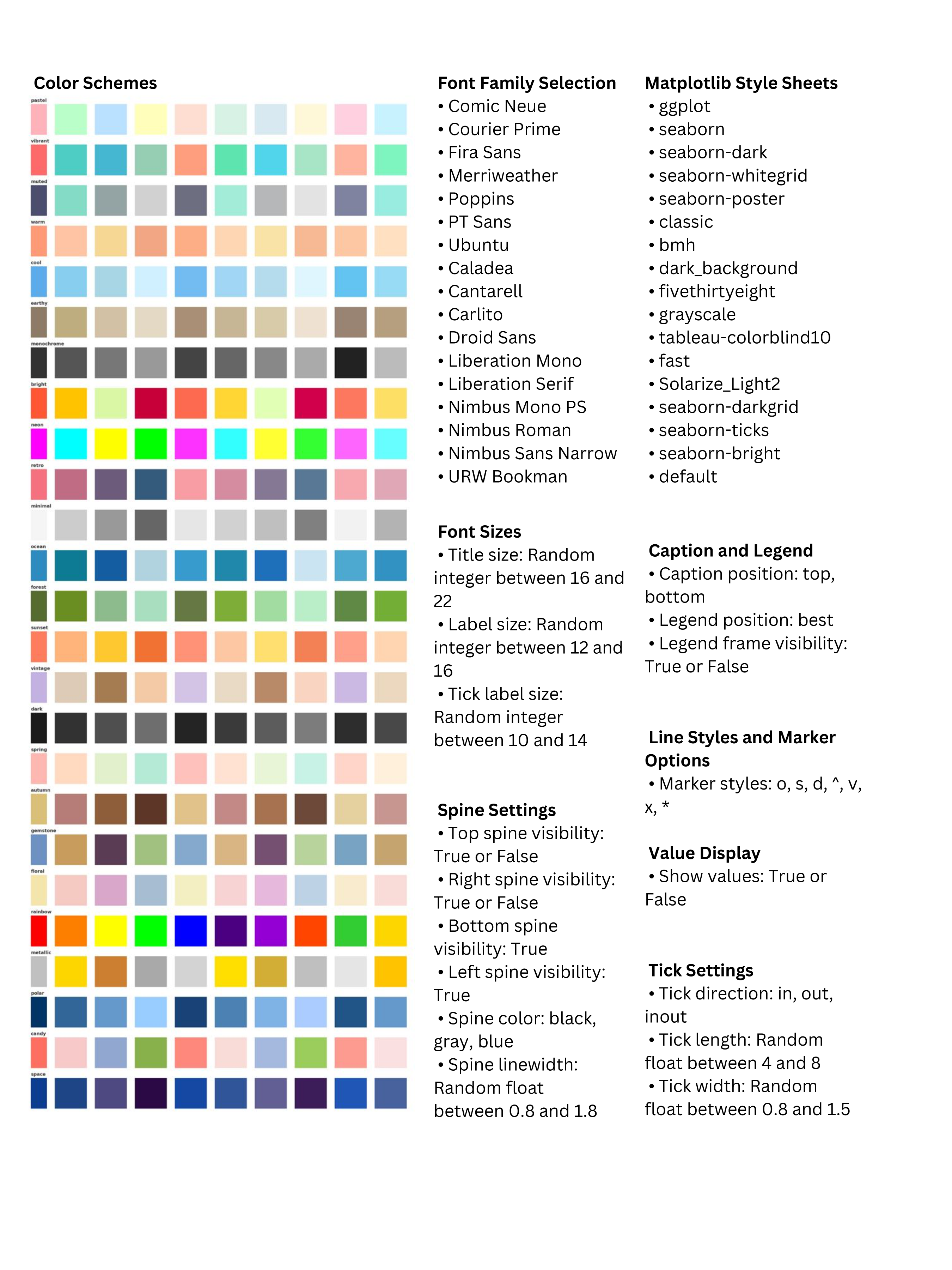}
    \caption{The design decision option space for MATSA synthetic charts, illustrating the various configurable elements and parameters available for customizing chart generation. This visual representation highlights the flexibility in chart design, encompassing aspects such as chart type, data presentation styles, and visual encoding options.}
    \label{fig:design-space}
\end{figure*}

Fig \ref{fig:design-space} represents the design decision space for MATSA synthetic charts. It illustrates the wide range of charts generated, showcasing the visual diversity, variations in layouts, and the impact of different design choices on the chart's structure and appearance.

\section{Algorithmic Heuristics for Point Extraction}

Algorithms \ref{alg:detect_bars} and \ref{alg:detect_pie_chart} depict the algorithmic workflow for mark identification and rendering, utilizing heuristics and SAM-based segmentation techniques. These algorithms effectively segment and label chart elements, enabling downstream fine-grained attribution.

\begin{algorithm}
\caption{Detect Bounding Boxes in Bar Charts}
\label{alg:detect_bars}
\begin{algorithmic}[1]
\Procedure{DetectBarBoundingBoxes}{image\_path, predictor}
    \State \textbf{Input:} Image path $image\_path$, predictor model $predictor$
    \State \textbf{Output:} Processed image, list of bounding boxes

    \State \textbf{Step 1: Preprocess Image}
    \State Load image and check validity
    \State Convert to grayscale and apply thresholding
    \State Perform morphological operations to clean noise

    \State \textbf{Step 2: Detect Initial Contours}
    \State Find contours in the thresholded image
    \State Filter contours by area to identify bar-like shapes

    \State \textbf{Step 3: Process Bar Contours}
    \For{each bar contour}
        \State Expand bounding box for analysis
        \State Mask region and extract unique colors
        \For{each unique color}
            \State Create mask and detect sub-contours
            \If{contour is rectangular}
                \State Store bounding box
            \EndIf
        \EndFor
    \EndFor

    \State \textbf{Step 4: Refine Bounding Boxes}
    \State Remove overlapping boxes and sort by position
    \For{each box}
        \State Use $SAM predictor$ to refine boxs
        \If{valid contour found in mask}
            \State Add final bounding box
        \EndIf
    \EndFor

    \State \textbf{Step 5: Finalize Output}
    \State Label bounding boxes and draw on image
    \State \textbf{Return:} Processed image, final list of bounding boxes
\EndProcedure
\end{algorithmic}
\end{algorithm}

\begin{algorithm}
\caption{Detect and Label Pie Chart Sectors}
\label{alg:detect_pie_chart}
\begin{algorithmic}[1]
\Procedure{DetectPieChartSectors}{image\_path, predictor}
    \State \textbf{Input:} Image path $image\_path$
    \State \textbf{Output:} Processed image with labeled pie chart sectors

    \State \textbf{Step 1: Preprocess Image}
    \State Load the image and convert it to grayscale
    \State Apply binary thresholding with Otsu's method
    \State Detect external contours and find the largest one

    \State \textbf{Step 2: Identify Center and Radius}
    \State Compute the minimum enclosing circle of the largest contour
    \State Extract the center $(x, y)$ and radius $r$

    \State \textbf{Step 3: Extract Pie Chart Region}
    \State Create a circular mask based on the detected center and radius and isolate the pie chart region

    \State \textbf{Step 4: Unroll the Pie Chart}
    \State Define sampling angles $\theta$ in $[0, 2\pi]$
    \State Sample pixel intensities along concentric ellipses at varying radii
    \State Store the unrolled intensities as a 2D array

    \State \textbf{Step 5: Detect Sector Boundaries}
    \State Apply the Sobel operator to detect horizontal edges in the unrolled image
    \State Compute a binary edge map by thresholding strong edges
    \State Identify complete edges that span most of the unrolled height
    \State Map these edges back to angles in $[0, 2\pi]$

    \State \textbf{Step 6: Process and  Refine  Sectors}
    \For{each candidate sector}
        \State Use $SAM predictor$ to refine sector
        \If{valid sector found in mask}
            \State Add final sector
        \EndIf
    \EndFor
    \State Compute midpoint angles for labeling sectors

    \State \textbf{Step 7: Finalize Output}
    \For{each sector}
        \State Label sectors on chart image
    \EndFor

    \State \textbf{Return:} Processed image with labeled sectors, final list of bounding boxes
\EndProcedure
\end{algorithmic}
\end{algorithm}

\end{document}